\documentclass[letterpaper]{article} 
\usepackage[submission]{aaai2026}  
\usepackage{times}  
\usepackage{helvet}  
\usepackage{courier}  
\usepackage[hyphens]{url}  
\usepackage{graphicx} 
\usepackage{natbib}  
\usepackage{caption} 
\usepackage{algorithm}
\usepackage{algorithmic}
\usepackage{amsmath}
\usepackage{booktabs}
\usepackage{amsfonts}
\usepackage{color}
\usepackage{framed}
\usepackage{graphicx}
\usepackage{lscape}
\usepackage{multirow}

\usepackage{newfloat}
\usepackage{listings}
\DeclareCaptionStyle{ruled}{labelfont=normalfont,labelsep=colon,strut=off} 
\floatstyle{ruled}
\newfloat{listing}{tb}{lst}{}
\floatname{listing}{Listing}
\title{DanceChat: Large Language Model-Guided Music-to-Dance Generation}
\author {
    Qing Wang,
    Xiaohang Yang,
    Yilan Dong,
    Jiahao Yang,
    Naveen Raj Govindaraj, \\
    Gregory Slabaugh,
    Shanxin Yuan\footnotemark[1]
}
\affiliations {
    Queen Mary University of London\\
    \{qing.wang, xiaohang.yang, yilan.dong, jiahao.yang, n.govindaraj, g.slabaugh, shanxin.yuan\}@qmul.ac.uk
}

\usepackage{bibentry}

\begin{document}

\maketitle
\begin{abstract}
Music-to-dance generation aims to synthesize human dance motion conditioned on music input. Despite recent progress, significant challenges remain due to the semantic gap between music and dance motion, as music offers only abstract cues, but lacks explicit physical movement descriptions. The challenge is further amplified by the scarcity of paired music and dance data, which restricts the model’s ability to learn diverse dance patterns. These limitations highlight the need for additional semantic guidance beyond the musical signal.
In this paper, we propose \textbf{DanceChat}, a novel framework that leverages a Large Language Model (LLM) as a choreographer to generate high-level textual instructions from structured music descriptions. These instructions serve as semantic guidance to bridge the gap between music and motion. DanceChat integrates music, beat, and text features into a unified representation, and employs a diffusion-based motion generator trained with a proposed multi-modal alignment loss. Extensive experiments on AIST++ dataset show that DanceChat outperforms state-of-the-art methods both qualitatively and quantitatively.
\end{abstract}


\footnotetext[1]{Corresponding author.}
\section{Introduction} \label{sec:intro}

Dance has long served as a powerful medium of expression in cultures, playing a central role in rituals, social connections, and artistic activities \cite{carroll2008feeling}. In the digital age, dance flourishes on social media, allowing people to share creativity and connect globally. Recent advances in deep learning have enabled computational exploration of this art form, particularly through music-to-dance generation, where models automatically create dance movements synchronized to music. This technology has impactful applications in entertainment, enabling realistic virtual idols and 3D characters in virtual reality. Beyond entertainment, automated dance generation is valuable in fields like robotics, digital human, and physical rehabilitation, where it helps create lifelike movement simulations and supports human-computer interaction. 

\begin{figure}[t]
\centering
    \captionsetup{type=figure}
    \includegraphics[width=1.\linewidth]{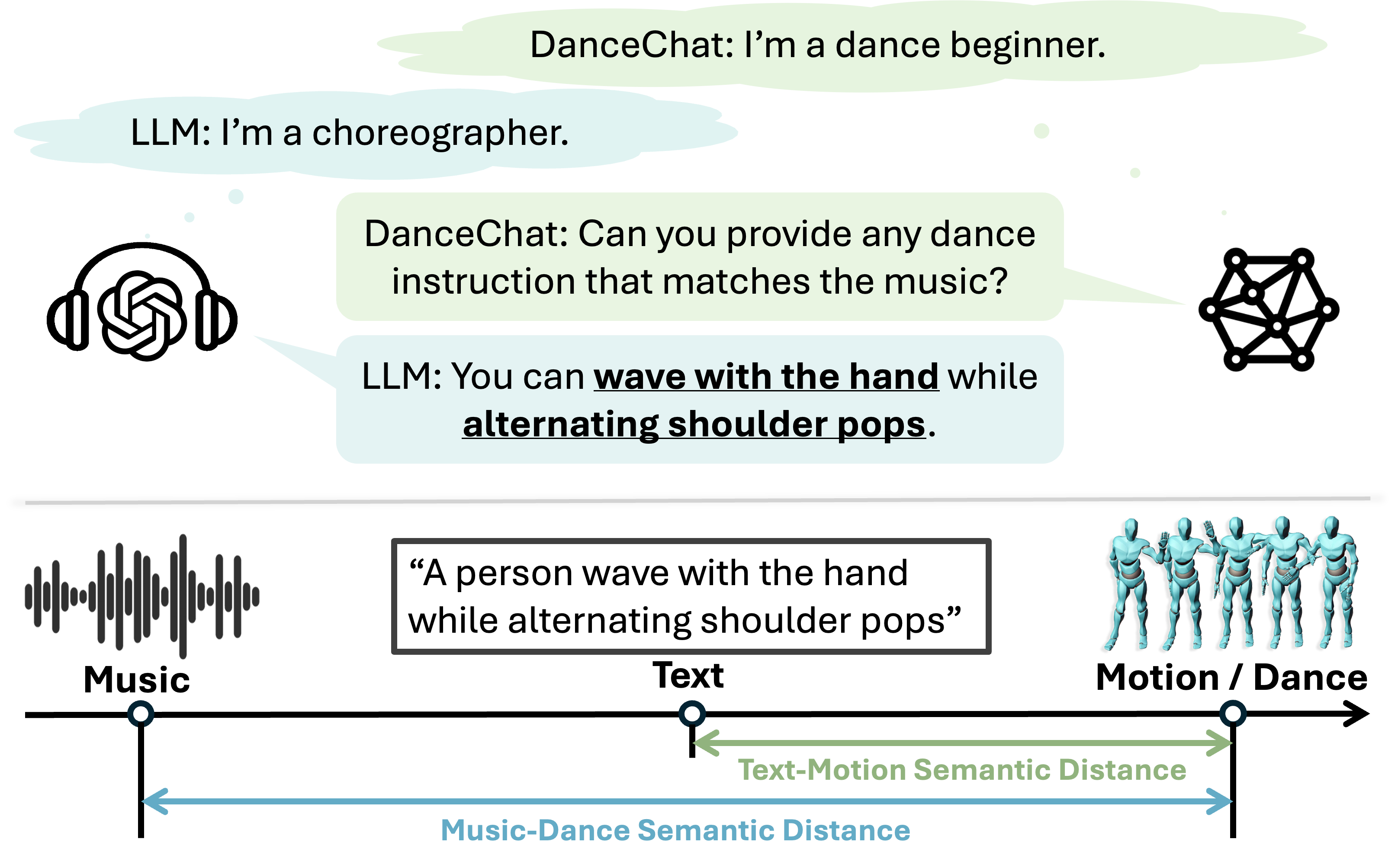}
    \captionof{figure}{Our approach draws inspiration from real-world dance learning, where learners rely on choreographers to interpret music. DanceChat leverages LLMs as pseudo-choreographers to translate music into textual instructions, bridging the semantic gap between music and motion. While the mapping from music to motion is inherently one-to-many and ill-posed, text offers a more structured and interpretable intermediary.}
    \vspace{-1em}
    \label{fig:idea}
\end{figure}

Most existing music-to-dance generation approaches fall into two main categories: regression models \cite{DBLP:conf/iclr/HuangHWSZJ21, li2021ai}, which are typically designed to directly learn a mapping from input music to dance motions, and generative models \cite{siyao2022bailando, tseng2023edge, zhang2024bidirectional, huang2024beat}, which aim to learn a distribution over plausible dances conditioned solely on the input music. However, both approaches lack explicit movement instructions, and rely only on the abstract nature of musical elements.
This creates a broad semantic gap between music and motion and complicates the task of generating physically plausible movements that align with the music’s genre and maintain aesthetic quality \cite{lee2019dancing}. Moreover, the scarcity of labeled dance data, due in part to the high cost of motion capture, makes it difficult for models to learn a well-generalized and diverse motion distribution. With limited training samples, the model struggles to capture the full variability of human dance, resulting in outputs that may lack stylistic diversity or fail to generalize to unseen music. 

These challenges highlight the need for advanced methods capable of bridging complex semantic gaps and injecting semantic-rich textual instructions without relying on additional annotations. Recent advancements in large language models (LLMs) have demonstrated remarkable success in capturing semantic relationships across various domains \cite{wu2023next, han2023imagebind, doh2024enriching, gardner2023llark, ding2024songcomposer}. 
They are trained on diverse content, including detailed discussions on music and dance, enabling them to interpret musical elements such as tempo, key, and chord progressions, as well as recognize movement styles, choreographic vocabulary, and dance semantics. This allows them to generate contextually appropriate descriptions that capture not only the structure and rhythm of a musical piece, but also align with plausible dance patterns. This inspires us to explore the potential of leveraging LLMs to enhance the diversity of music-to-dance generation. 

Building on these insights, our method adapts the capabilities of LLMs to address the semantic gap in music-driven dance generation. We propose a novel approach, named \emph{DanceChat}, which conceptualizes the generation model as a \textit{dancing learner} and leverages LLMs as a \textit{pseudo-choreographer} (Fig. \ref{fig:idea}). Specifically, we extract structured musical descriptors, including key, tempo, and chord progressions alongside the dance genre context, and feed this information into LLM, which then generates corresponding textual dance instructions, serving as a pseudo-guide to condition the motion generation model during training. This helps reduce the semantic gap by offering interpretative cues that map abstract musical elements to concrete motion patterns.

To further enhance cross-modal alignment, we explicitly extract beat and downbeat information from the input music as an additional rhythmic modality, which is known to correlate strongly with movement timing in dance. The original music, rhythmic cues, and LLM-generated textual instructions are then integrated through a multi-modal fusion module, enabling the model to form a richer and more coherent understanding of the music. By combining music, beat signals, and semantic instructions, DanceChat produces dance motions that are rhythmically aligned, physically plausible, and stylistically expressive.

In summary, our contributions are three-fold:


1) A novel LLM-guided framework that generates context-driven dance movement instructions aligned with specific musical structure and style.


2) A multi-modal fusion paradigm that integrates beat-aware musical information with the textual dance instructions, enriching the model’s capacity for music-to-dance alignment.

3) A multi-modal alignment loss that leverages text as an intermediary to enhance the semantic connection between music and dance, reducing the gap between these modalities and improving the coherence and expressiveness of generated dance movements.

\section{Related Works}

\label{sec:related_work}

\begin{figure*}[ht]
  \centering
   \includegraphics[width=1.0\linewidth]{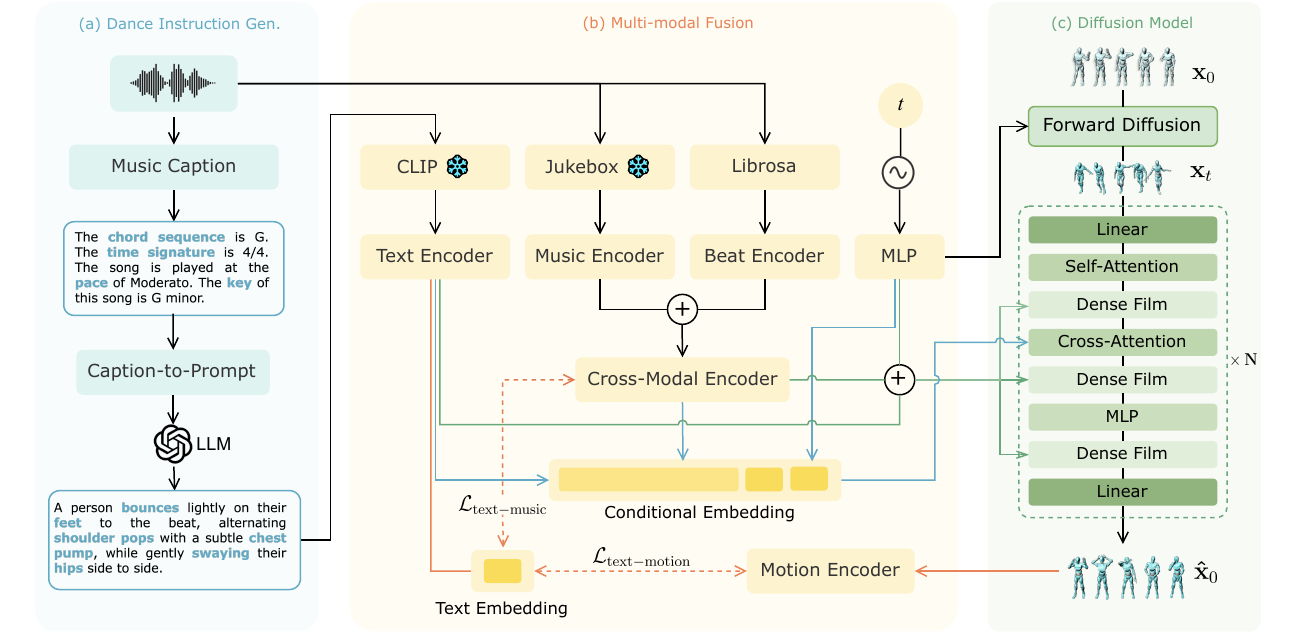}

   \caption{Overview of our approach. DanceChat consists of three main components: (a) Dance Instruction Generation module extracts music caption (tempo, key, chord, etc.) from the given music condition, (b) Multi-modal Fusion module integrates the encoded representations of music, beats, and text into a unified conditional embedding, and (c) Diffusion-based Motion Synthesis module uses this unified embedding to guide the generation of realistic dance movement.}
   \vspace{-1em}
   \label{fig:overview}
\end{figure*}

\subsection{3D Human Motion Generation}

3D human motion generation has become a central problem in computer vision and graphics, with applications in animation, robotics, and virtual agents. Recent research has explored a wide range of conditioning modalities, including natural language, audio, scene layout, and control signals, aiming to improve realism, diversity, and controllability. Among these, text-conditioned motion generation has emerged as a promising direction, leveraging the semantic clarity and expressiveness of natural language to provide interpretable and flexible guidance. Diffusion-based approaches \cite{kim2023flame, zhang2024motiondiffuse, liang2024intergen, zhou2024emdm} have demonstrated that structured textual inputs can guide the synthesis of diverse and coherent motion. This highlights the potential of language as a strong semantic prior for motion planning. Building on this idea, our work proposes to use language as a semantic bridge between music and motion. By translating musical attributes into textual instructions, we enable explicit, high-level guidance for music-to-dance generation, addressing the limitations of relying solely on music features.

\subsection{Music-to-Dance Generation} 

The Music-to-Dance generation task poses unique challenges due to the need for models to understand and generate dance sequences that are both rhythmically and stylistically aligned with music. 
Previous methods \cite{chan2019everybody, zhuang2022music2dance, li2023finedance, tseng2023edge, yang2023longdancediff, tseng2023edge, huang2024beat} incorporate advanced audio-processing techniques with motion generation algorithms to enhance the flexibility and expressiveness of generated dances. 
Despite these advancements, existing approaches still face limitations in bridging the substantial semantic gap between music and dance, making the synchronization of dance sequences a significant challenge.
Unlike text-to-motion generation where textual cues explicitly describe body movements, music provides only implicit cues through rhythm, tempo, and chord, which do not directly correspond to physical actions. This gap complicates the creation of dance sequences that are both well-aligned with the music and stylistically diverse. 

\subsection{Large Language Models}
Recent instruction-tuned LLMs \cite{openai2023gpt4, anthropic2023claude, touvron2023llama, deepseek2024} have demonstrated strong capabilities in capturing contextual semantics, following user intent, and generalizing across domains. 
Inspired by these advancements, researchers have begun exploring LLMs for non-linguistic tasks, including motion generation \cite{zhou2024avatargpt, ribeiro2024motiongpt}, where their ability to model temporal coherence and semantic structure enables fine-grained motion control.
To effectively integrate LLMs into non-linguistic domains, prompt engineering has emerged as a key technique \cite{liu2021makes, liu2022design, oppenlaender2023taxonomy}. This involves crafting structured prompts that effectively communicate with models, directing the generation toward desired outcomes, and allowing users to control both the content and style of the outputs. In motion generation, prompt engineering enables models to translate high-level textual instructions into semantically coherent motion sequences, enhancing controllability and expressiveness \cite{motionagent}. Building upon these developments, our work introduces a novel approach that leverages LLM-generated textual choreography instructions to bridge the semantic gap between music and dance.

\section{Our Approach}
\label{sec:our_approach}

\subsection{Problem Definition}
The task of music-to-dance generation involves producing a dance motion sequence $\mathcal{X} = \{\mathit{x}^1, \mathit{x}^2, \dots, \mathit{x}^N \}$ from a given music track $\mathcal{M}$ with a duration of $\mathit{N}$ frames. Here, $\mathit{x}^{i} \in \mathbb{R}^D$ represents a human pose at the \textit{i}-th frame, which is denoted as a \textit{D}-dimensional vector. We present the dance sequences with 24-joint SMPL format \cite{loper2023smpl}, which represents each joint with 6-DOF rotation representation \cite{zhou2019continuity} and a single root translation. Along with the binary contact labels $b\in \{0,1\}^{2\times 2=4}$ for the heel and the toe of each foot, the dimension $D=24\times6 +3+4 = 151$.

\subsection{Overview}

Our model architecture is illustrated in Fig. \ref{fig:overview} and consists of three main components: \textit{Dance Instruction Generation}, \textit{Multi-modal  Feature Extraction and Fusion}, and \textit{Diffusion-based Motion Synthesis}. The \textit{Dance Instruction Generation} module (Fig. \ref{fig:overview}a) extracts music caption (tempo, beat, key, chord) from the given original music  $\mathcal{M}$ and utilizes an LLM as a pseudo-choreographer to transform this information into dance motion instructions $\mathcal{T}$, providing textual guidance for motion generation. 
The \textit{Multi-modal Fusion} module (Fig. \ref{fig:overview}b) then integrates the encoded representations of music $\mathcal{M}$, beats $\mathcal{B}$, and text $\mathcal{T}$ into a unified conditional embedding. This fused representation aligns music, beat, and movement instructions, creating a cohesive context for dance generation.
Finally, the \textit{Diffusion-based Motion Synthesis} module (Fig. \ref{fig:overview}c) uses this unified embedding to guide the generation of dance movements that capture the style and dynamics of the music.

\subsection{Dance Instruction Generation}

This module is responsible for transforming music into high-level textual dance instructions, which serve as semantic guidance for the dance generation process. This module comprises two key steps: Music Caption Construction and LLM-based Instruction Generation. 

\noindent\textbf{Music Caption Construction.} 
We adopt a structured captioning procedure to extract and represent key musical attributes from each input music clip. Building upon the framework proposed by Mustango \cite{melechovsky2023mustango}, we extract four common symbolic descriptors: tempo, chords, keys, and beat structure using a set of specialized tools. To further enhance clarity and generation consistency, we follow a controlled template-based sentence construction process, converting each musical attribute into interpretable textual music captions (e.g., `The chord sequence is G. The time signature is 4/4. The song is played at the pace of Moderato. The key of this song is G minor.'). These captions are appended to the input prompts to provide rich meta-level guidance for downstream instruction generation.

\begin{figure}[]
\centering
    \captionsetup{type=figure}
    \includegraphics[width=1.\linewidth]{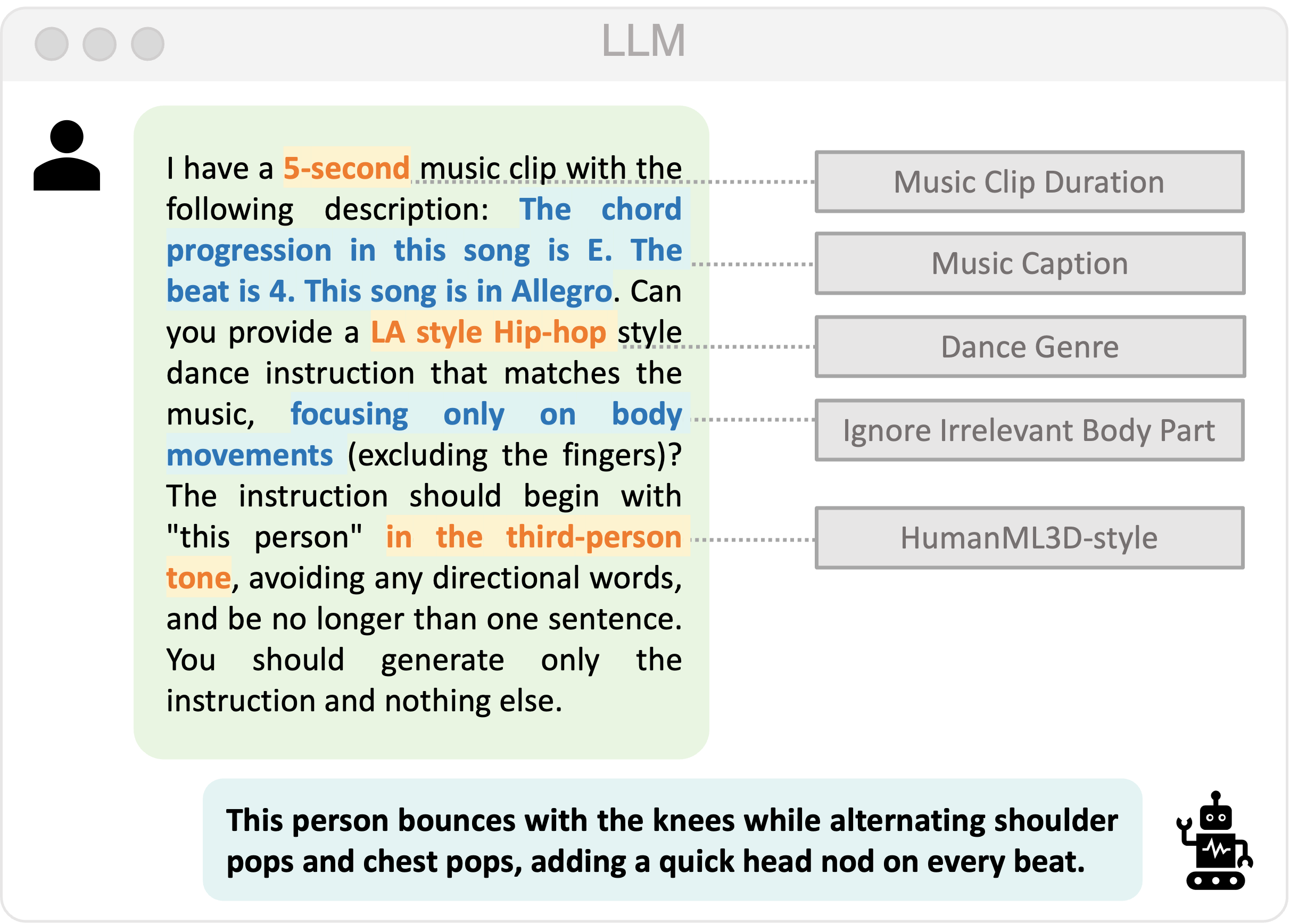}
    \captionof{figure}{Details of our LLM-based textual instruction generation process. The prompt guides an LLM to generate HumanML3D-style dance instructions.}
    \vspace{-1em}
    \label{fig:prompt}
\end{figure}

\noindent\textbf{LLM-based Instruction Generation.} 
To convert music into interpretable guidance for motion, we generate a textual dance instruction $\mathcal{T}$ from the music caption using an instruction-following LLM. The LLM acts as a pseudo choreographer, producing stylistically and rhythmically aligned instructions for each music clip. These instructions provide explicit descriptions of body movements, offering fine-grained guidance to the motion generator.
To ensure consistency and controllability, we adopt a carefully designed prompt engineering strategy inspired by the format and motion description style of the HumanML3D dataset \cite{guo2022generating}. As shown in Fig \ref{fig:prompt}, we incorporate key elements such as clip duration and dance style. We also avoid directional words and overly fine-grained spatial cues to improve clarity and reduce ambiguity. 

\noindent\textbf{Instruction Results.} 
The output of this process is a set of natural language dance instructions that provide semantic cues for the dance generation model. For instance (Fig. \ref{fig:prompt}), given a music track with a moderate tempo and a Hip-Hop style, the generated instruction might read: \textit{This person bounces with the knees while alternating shoulder pops and chest pops, adding a quick head nod on every beat.} These instructions capture both the rhythmic alignment and stylistic characteristics of the music, bridging the semantic gap between music and motion.

\subsection{Multi-modal Feature Extraction and Fusion}
\begin{table*}[ht]
\centering
\renewcommand{\arraystretch}{1.2}
\setlength{\tabcolsep}{3mm}{
\begin{tabular}{lcccc}
\toprule[0.5mm]
Method & PFC $\downarrow$  & $\mathrm{Div}_{k} \rightarrow$  & $\mathrm{Div}_{g} \rightarrow$ & Beat Align. Score $\uparrow$ \\ \hline
Ground Truth                & 1.332 & 10.61 & 7.48 & 0.24             \\ \hline
FACT \cite{li2021ai}         & 2.254 & 10.85 & 6.14 & 0.22               \\
Bailando \cite{siyao2022bailando} & 1.754 & 7.92 & 7.72 & 0.23          \\
EDGE \cite{tseng2023edge}    & 1.536 & 9.48  & 5.72  & 0.23               \\
BADM \cite{zhang2024bidirectional}  & 1.424 & 8.29  & 6.76 & 0.24               \\
Beat-it \cite{huang2024beat} & 0.966 & 9.66 & 7.25 & \textbf{0.66} \\ \hline
Ours DanceChat & \textbf{0.828} & \textbf{10.64} & \textbf{7.36} & \underline{0.27} \\ 
\bottomrule[0.5mm]
\end{tabular}}
\caption{Quantitative comparison with state-of-the-art methods on the AIST++ dataset. FACT and Bailando are Transformer- and VQ-VAE-based models that require a starting pose. EDGE, BADM, and Beat-it are diffusion-based approaches. $\uparrow$ means higher is better, $\downarrow$ means lower is better, $\rightarrow$ means closer to ground truth is better. \textbf{Bold} indicated the best result; \underline{underline} indicates the second-best. Our method outperforms the other methods on $\mathrm{PFC}$, $\mathrm{DIV}_k$ and $\mathrm{DIV}_g$, and performs the second-best on $\mathrm{BAS}$.}
\vspace{-1em}
\label{tab: quantitative}
\end{table*}
This module integrates multiple conditions from text, music, and beat information into a unified conditional embedding. We first encode each modality individually, then hierarchically fuse these embeddings to leverage their complementary strengths effectively.

\subsubsection{Multi-modal Feature Extraction}
\
\newline
\noindent\textbf{Music Encoding.} We leverage a pre-trained Jukebox \cite{dhariwal2020jukebox} model to extract high-level musical features from raw audio inputs. Specifically, we input the raw audio into the frozen Jukebox encoder, obtaining embeddings that represent the musical content. To adapt these embeddings for our task, we introduce a trainable two-layer Transformer encoder that processes the Jukebox-derived features, allowing the model to learn task-specific representations while retaining the rich musical information captured by Jukebox, resulting in embeddings $E_M\in \mathbb{R}^{d_M}$.

\noindent\textbf{Text Encoding.} 
We process the LLM-generated dance instructions using a CLIP-based text encoder \cite{radford2021learning} followed by a trainable Transformer to extract task-specific embeddings. To enhance semantic alignment, we initialize the encoder with pretrained weights from MotionDiffuse \cite{zhang2024motiondiffuse}, which is trained on HumanML3D-style \cite{guo2022generating} descriptions. This setup enables our model to interpret dance instructions effectively, without requiring paired text-motion supervision. The final textual embedding is denoted as $E_T\in \mathbb{R}^{d_T}$.


\noindent\textbf{Beat Encoding.} 
To enhance rhythmic awareness in generated motion, we explicitly incorporate beat proximity features. We use Librosa~\cite{mcfee2015librosa} to extract beat and downbeat positions from audio. These discrete time points mark the perceived rhythmic structure of music. To convert this into a continuous representation, we compute a Gaussian-shaped pulse for each frame centered around the nearest beat/downbeat. This reflects the frame’s temporal distance to nearby rhythmic accents, enabling the model to perceive and respond to rhythm over time. We then project these proximity values into a learnable embedding space using a linear projection layer, resulting in beat embeddings $E_B\in \mathbb{R}^{d_B}$. 

Together, the extracted features from all modalities form a unified representation: the textual embedding $E_T$ provides pseudo motion guidance, the musical embedding $E_M$ captures style and structure, and the beat embedding $E_B$ encodes rhythmic timing, enabling effective and complementary cross-modal integration.

\subsubsection{Multi-modal Hierarchical Fusion}
\
\newline
We fuse these modality embeddings through a hierarchical, two-level strategy designed to incrementally integrate rhythmic, musical, and textual guidance.

\noindent\textbf{Music-Beat Rhythmic Fusion.} Initially, we fuse music embeddings $E_M$ with rhythmic beat embedding $E_B$ using an additive fusion approach:
\begin{equation}
    E_M^\prime = E_M + E_B,
\end{equation}
This straightforward yet effective fusion explicitly integrates rhythmic timing information directly into the music representations. By enhancing music embeddings with beat cues, it enables the model to accurately capture rhythmically significant moments within the music.

\noindent\textbf{Music-Text Semantic Fusion.}
Subsequently, we incorporate textual information from dance instruction embeddings $E_T$ to provide semantic guidance for dance generation. Specifically, we concatenate the textual embeddings $E_T$ with the previously fused musical-beat embeddings $E_M^\prime$:
\begin{equation}
    E_F = [E_M^\prime; E_T],
\end{equation}
generating the final unified conditional embedding $E_F$. This second fusion stage integrates high-level choreographic instructions, derived from natural language guidance, with detailed rhythmic-musical context. Thus, the final embedding $E_F$ effectively combines semantic choreographic concepts from textual instructions and detailed rhythmic structure from music and beat embeddings.

The unified embedding $E_F$ is then used to condition the diffusion-based motion synthesis module, guiding the generation of realistic and expressive dance that reflects both musical structure and textual intent.

\subsection{Diffusion-based Motion Synthesis}
\subsubsection{Diffusion Model}
\
\newline
This module leverages a denoising diffusion probabilistic model (DDPM) \cite{ho2020denoising} to progressively guide noisy samples toward realistic dance motions, conditioned on the multi-modal embedding  $E_F$  from the previous Multi-modal Fusion stage.

We model dance generation as a Markovian diffusion process. The forward diffusion process begins with a clean motion sequence $\mathbf{x}_0$ and iteratively adds Gaussian noise over $T$ timesteps, eventually transforming it into a nearly random noise sample  $\mathbf{z}_T$. Formally, at each timestep $t$, the forward process is defined as:
\begin{equation}
  \mathit{q}(\mathbf{z}_t|\mathbf{z}_{t-1}) = \mathcal{N}(\sqrt{\mathit{1-\beta_t}}\mathbf{z}_{t-1}, \beta_t\mathit{I}),
  \label{eq:diffution_forward}
\end{equation}
where $\beta_t\in(0,1)$ is a variance schedule parameter that controls the amount of noise added at each step. By the final timestep $T$, the noisy sequence $\mathbf{z}_T$ closely approximates a standard Gaussian distribution $\mathcal{N}(0, \mathit{I})$.

To recover the original motion from this noisy representation, we employ a reverse diffusion process. This reverse process begins with the noisy sample $\mathbf{z}_T$ and iteratively denoises it, guided by the conditional embedding $E_F$, to approximate the original sequence $\mathbf{x}_0$. 
We adopt the simplified training objective from DDPM \cite{ho2020denoising}, minimizing the mean squared error (MSE) between the model's predictions and the true motion data across all timesteps. 
Formally, the training objective is defined as:
\begin{equation}
  \mathcal{L}_{\mathrm{simple}} = 
  \mathbb{E}_{\mathbf{x},t}[\parallel\mathbf{x}_0 - \mathbf{\hat{x}}_{\theta}(\mathbf{z}_t, t, E_F)\parallel^{2}_{2}],
  \label{eq:diffution_optimization}
\end{equation}
where $\mathbf{\hat{x}}_{\theta}(\mathbf{z}_t, t, E_F)$ denotes the reconstructed motion at timestep $t$ conditioned on the multi-modal embedding $E_F$ with model parameters $\theta$.
\subsubsection{Loss Functions}
\ 
\newline 
\noindent\textbf{Kinematic Loss.} To enhance the physical plausibility of generated motions, we incorporate several auxiliary losses commonly used in kinematic motion generation. In addition to the basic reconstruction loss $\mathcal{L}_{\mathrm{simple}}$, we adopt auxiliary terms introduced by Tevet et al.\cite{tevet2022human} and further refined in EDGE\cite{tseng2023edge}, targeting three key aspects of motion realism: joint positions (Eq. \ref{eq:loss_joint}), joint velocities (Eq. \ref{eq:loss_vel}), and foot-ground contact consistency (Eq. \ref{eq:loss_contact}). Below, $\mathbf{FK(\cdot)}$ denotes the forward kinematic function that converts joint angles into joint positions and the $\mathit{i}$-th superscript denotes the frame index. We note Eq. \ref{eq:loss_contact} only applies to the foot joints, and $\mathbf{\hat{b}}_i$ is binary foot contact signal.
\begin{equation}
  \mathcal{L}_{\mathrm{joint}} = 
  \frac{1}{\mathit{N}}
  \sum^{\mathit{N}}_{i=1}
  \parallel\mathbf{FK}(\mathbf{x}_i) - \mathbf{FK}(\mathbf{\hat{x}}_i)\parallel^{2}_{2}
  \label{eq:loss_joint}
\end{equation}

\begin{equation}
  \mathcal{L}_{\mathrm{vel}} = 
  \frac{1}{\mathit{N}-1}
  \sum^{\mathit{N}-1}_{i=1}
  \parallel(\mathbf{x}_{i+1}-\mathbf{x}_{i}) - (\mathbf{\hat{x}}_{i+1}-\mathbf{\hat{x}}_{i})\parallel^{2}_{2}
  \label{eq:loss_vel}
\end{equation}
\begin{equation}
  \mathcal{L}_{\mathrm{contact}} = 
  \frac{1}{\mathit{N}-1}
  \sum^{\mathit{N}-1}_{i=1}
  \parallel(\mathbf{FK}(\mathbf{x}_{i+1})-\mathbf{FK}(\mathbf{\hat{x}}_i))\cdot\mathbf{\hat{b}}_i\parallel^{2}_{2}
  \label{eq:loss_contact}
\end{equation}

\noindent\textbf{Multi-modal Alignment Loss.}
To bridge the semantic gap between music and motion, we design a multi-modal alignment loss that leverages text as an intermediate modality. Unlike prior methods that directly align music and motion, we instead calculates similarity between \textit{motion-text} and \textit{text-music} pairs at each timestep $\mathit{t}$ in the diffusion process. By using text as a shared semantic space, the model captures interpretable features related to both musical cues and motion dynamics. Specifically, for each timestep, we compute cosine similarity between motion-text embeddings and text-music embeddings, promoting alignment across all three modalities through the text-based intermediary. The total alignment loss is defined as follow: 

\begin{equation}
  \mathcal{L}_{\mathrm{align}} = 
  \frac{1}{\mathit{N}}
  \sum^{\mathit{N}}_{i=1}
  [(1-\mathrm{sim}(E_{t_i}, E_{m_i})) + (1-\mathrm{sim}(E_{t_i}, E_{x_i}))]
  \label{eq:loss_mm}
\end{equation}
This loss encourages consistent semantic alignment between music and motion via text.
It is combined with the other auxiliary losses, weighted by factors $\lambda$ to balance the magnitudes of the losses during training.

\begin{equation}
\begin{split}
  \mathcal{L} & = \mathcal{L}_{\mathrm{simple}} + \lambda_{\mathrm{joint}}\mathcal{L}_{\mathrm{joint}} + \lambda_{\mathrm{vel}}\mathcal{L}_{\mathrm{vel}} + \\
  &\lambda_{\mathrm{contact}}\mathcal{L}_{\mathrm{contact}} + \lambda_{\mathrm{align}}\mathcal{L}_{\mathrm{align}}
\end{split}
  \label{eq:loss_all}
\end{equation}

\section{Experiments}
\label{sec:experiments}

\subsection{Experiment Setup}

\begin{figure*}[ht]
  \centering
   \includegraphics[width=1.0\linewidth]{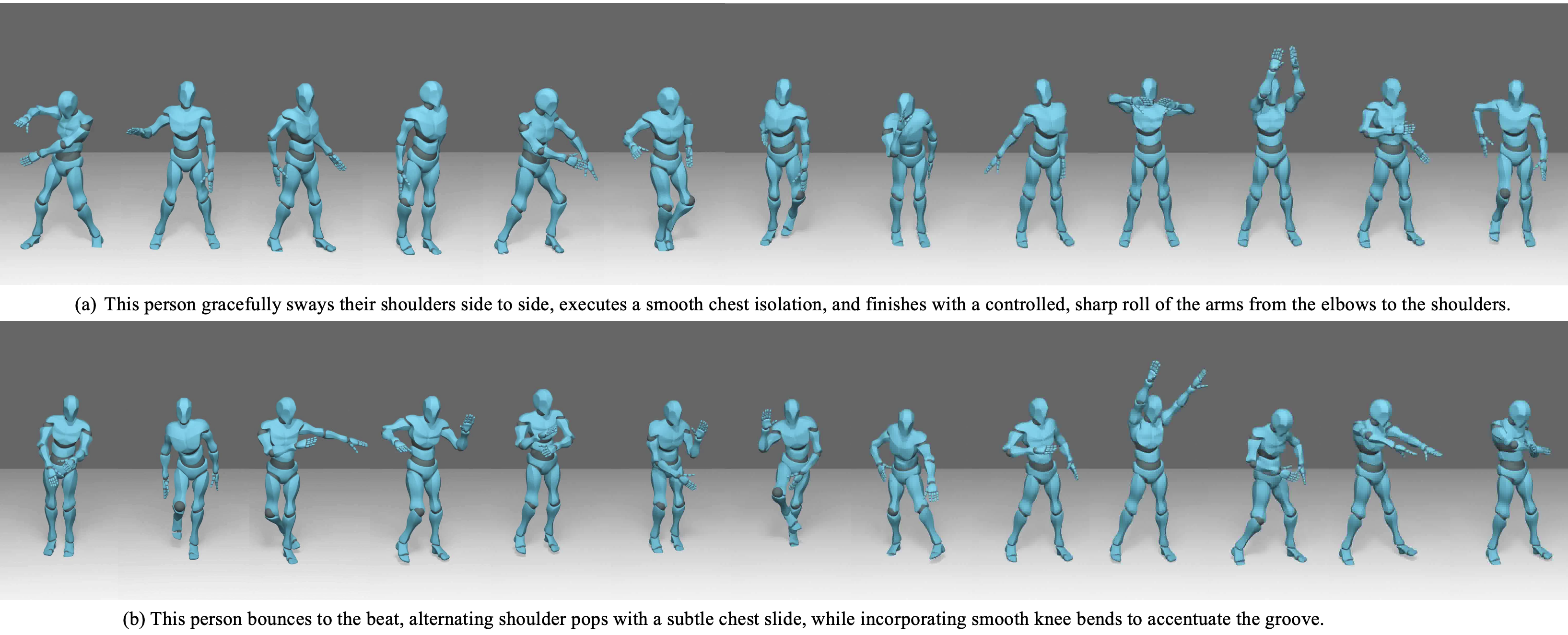}

   \caption{Example results generated by our model. Each sequence illustrates a 3D dance motion generated from a given music clip, conditioned on an LLM-generated textual instruction. The sentences below each motion sequence are produced by the LLM and serve as choreography prompts.}
   \label{fig:results}
\end{figure*}

\noindent\textbf{Dataset.}  We use the AIST++ \cite{li2021ai} dataset for both training and testing, which contains 1,408 music-dance pairs across 10 street dance genres. Each genre is paired with 6 distinct music tracks, enabling rhythmic and stylistic variation.
We follow the standard train/test split and segment the sequences into 5-second clips at 30 FPS. Music captions and textual guidance are generated based on each segmented clip.




\noindent \textbf{Implementation details.} In the dance instruction generation module, we utilize the ChatGPT-4o API \cite{chatgptapi} to generate textual dance motion instruction based on pre-extracted music captions. 
For music encoding, we extract features using a frozen Jukebox \cite{dhariwal2020jukebox} model and subsequently apply two additional Transformer encoder layers to refine the temporal and semantic representations. For text encoding, we use the frozen text encoder from CLIP ViT-B/32 \cite{radford2021learning} followed by another 4-layer Transformer encoder to adapt the output for conditional guidance. The latent dimensionality of both the music and text encoders is set to 512. 
Our model has 54M total parameters, and was trained on 1 NVIDIA A100 GPU for 48 hours with a batch size of 128.

\subsection{Comparison to Existing Methods}
\begin{table}
\centering
\begin{tabular}{lcc}
\toprule[0.5mm]
Method  & Ours (AIST++)  & Ours (In-the-Wild)    \\
\midrule
FACT                 & 86.25\%   & N/A         \\
Bailando             & 61.88\%   & N/A         \\
EDGE                 & 60.00\%   & 66.36\%     \\
\midrule
Ground Truth         & 48.12\%   & N/A         \\
\bottomrule[0.5mm]
\end{tabular}
\caption{User study results on AIST++ test set and in-the-wild music. We present the Preference Rate $\uparrow$ of user evaluations that prefer DanceChat (Ours) to FACT \cite{li2021ai}, Bailando \cite{siyao2022bailando}, and EDGE \cite{tseng2023edge}.}
\label{tab:user_study}
\end{table}
We have selected FACT \cite{li2021ai}, Bailando \cite{siyao2022bailando}, EDGE \cite{tseng2023edge}, BADM \cite{zhang2024bidirectional} and Beat-it \cite{huang2024beat} for comparisons with our DanceChat method. FACT applies a full-attention mechanism for cross-modal dependency modeling, using autoregressive generation for temporal consistency. Bailando enhances this with a VQ-VAE \cite{van2017neural} and GPT \cite{radford2018improving}-based framework, together with actor-critic reinforcement learning. EDGE adopts a transformer-based diffusion model, which serves as the backbone of our method. BADM builds on EDGE by proposing a bidirectional autoregressive diffusion model for smoother local motion. Beat-it further extends EDGE by integrating beat supervision and key-pose guidance to improve rhythm-movement alignment.


To evaluate the effectiveness of our approach, we conduct a series of experiments to assess the performance in terms of physical plausibility, motion diversity, and beat synchronization. The quantitative metrics are shown in Tab. \ref{tab: quantitative}. In addition to this, we also conduct a user study to gather subjective feedback on the overall quality and coherence of the generated motions, as shown in Tab. \ref{tab:user_study}.

\noindent\textbf{Motion Quality.} 
We use the Physical Foot Contact (PFC) score \cite{tseng2023edge} to evaluate foot-ground interactions, a key factor for physical realism. It measures the accuracy of foot contact, ensuring that generated sequences avoid unrealistic foot sliding and maintain natural center of mass (COM) dynamics. Lower scores indicate more stable and realistic foot-ground interactions. As shown in Tab. \ref{tab: quantitative}, DanceChat achieves the best PFC performance, improving by 0.138 over SOTA method Beat-it. This highlights our model’s ability to produce physically plausible dance sequences.

\noindent\textbf{Motion Diversity.}
To assess diversity, we compute the mean Euclidean distance of motion features across 5-second clips, following standard metrics $\mathrm{Div}_{k}$ (kinematic space) and $\mathrm{Div}_{g}$ (geometric space) outlined by Bailando. Our metrics are consistent with prior work and computed to match the distributional spread of the ground truth, avoiding the inflation from jittery or unstable motion. As shown in Tab. \ref{tab: quantitative}, our approach achieves a $\mathrm{Div}_{g}$ of 7.36 and $\mathrm{Div}_{k}$ of 10.64, both closest to ground truth among all methods, reflecting a good balance between variation and realism.


\noindent\textbf{Beat Alignment Score (BAS).}
We use the BAS metric to evaluate the rhythmic alignment of generated dance with music, which measures the consistency between musical beats and joint speed minima in motion. As shown in Tab. \ref{tab: quantitative}, our model achieves the second-best BAS score, demonstrating strong beat-tracking ability. While Beat-it, which uses beat-supervised training, achieves the highest score, our method obtains competitive performance without explicit beat-level supervision. This is enabled by our flexible multi-modal fusion framework, which jointly encodes music and beat features to produce rhythm-aware motion and generalize across styles.

\noindent\textbf{User Study.}
To assess the visual and rhythmic quality of our method, we conduct a user study comparing DanceChat with baseline methods on AIST++ and in-the-wild music. 30 participants each evaluated 20 pairs of 10s videos, where each pair included a DanceChat result and one from a competing method, synchronized to the same music. Participants were asked to select the more appealing video. Results are shown in Tab. \ref{tab:user_study}. DanceChat outperforms all baselines on AIST++. In the in-the-wild setting, it achieves a 66.36\% preference over EDGE, confirming its robustness to unseen music. Additionally, an observable reduction in foot sliding is noted, further confirming the PFC improvement.

\subsection{Ablation Study}
We perform an ablation study on the AIST++ to evaluate the impact of different modalities and the effectiveness of our multi-modal alignment loss. Results in Tab. \ref{tab:ablation} are reported using $\mathrm{PFC}$, $\mathrm{DIV}_k$, and BAS, where lower $\mathrm{PFC}$ and higher others indicate better performance.

\noindent\textbf{Impact of Modality Inputs.}
To understand the individual contributions of each modality, we assess various configurations by selectively including or excluding the music (M), beat (B), and text (T) modalities. 
The baseline model, using only music (M), performs the worst, showing limited diversity and weak rhythmic alignment. Adding beat information (M+B) enhances rhythmic alignment, indicating that beat cues contribute both to physical realism in generated movement patterns and to synchronization with the music.
Adding textual guidance (M+T) leads to the best performance in PFC and diversity, suggesting that text improves motion realism and expressiveness by providing explicit semantic choreography cues. However, lacking beat information, this configuration underperforms in BAS.
Combining all three modalities (M+B+T) produces the highest BAS and consistently strong results in other metrics.

\noindent\textbf{Impact of Multi-Modal Alignment Loss.}
We further incorporate the multi-modal alignment loss (MM Align. Loss), which leverages text as an intermediary bridge between music and motion to narrow the semantic gap between these modalities. 
The alignment loss improves $\mathrm{PFC}$) and $\mathrm{DIV}_k$, demonstrating its effectiveness in  cross-modal integration. This suggests that using text as a bridge in the alignment loss helps create semantically rich and rhythmically aligned dance motions that benefit from the strengths of each modality without compromising diversity or quality.



\begin{table}
\centering
\begin{tabular}{ccccccc}
\toprule[0.5mm]
\multicolumn{3}{c}{Ablations} & & \multicolumn{3}{c}{Metrics} \\ 
\cline{1-3} \cline{5-7}
M & B & T && PFC $\downarrow$ & Div$_k$ $\rightarrow$ & BAS $\uparrow$ \\
\midrule

\checkmark  & \space        & \space     && 1.536 & 9.48 & 0.23 \\
\checkmark  & \checkmark    & \space     && 1.066 & 8.76 & 0.27 \\
\checkmark  & \space        & \checkmark && 1.166 & 9.69 & 0.26 \\
\checkmark  & \checkmark    & \checkmark && \underline{0.843} & \underline{9.85} & \textbf{0.28} \\
\multicolumn{3}{c}{MM Align. Loss} && \textbf{0.828} & \textbf{10.64} & \underline{0.27} \\
\midrule
\multicolumn{3}{c}{Ground Truth} && 
1.332 & 10.61 & 0.24 \\
\bottomrule[0.5mm]
\end{tabular}
\caption{Ablation study of modalities, where M indicates music, B indicates beat and T indicates text. The best results are indicated in \textbf{bold}, and the second best are \underline{underlined}.}
\label{tab:ablation}
\end{table}

\section{Conclusion}
\label{sec:discussion_and_conclusion}
We present DanceChat, a novel multi-modal framework that leverages language as an intermediate modality to bridge the semantic gap between music and dance. By incorporating textual dance instructions generated by LLMs and integrating them with both music and beat features, our approach enables semantically guided and rhythmically aligned motion generation. 
Comprehensive experiments on AIST++ dataset demonstrate that DanceChat achieves strong performance across multiple aspects of dance quality, including physical plausibility, motion diversity, and beat synchronization. Ablation results confirm that each modality contributes uniquely to dance quality, while our multi-modal alignment loss plays a crucial role in harmonizing these modalities to produce musically aligned motion sequences. Beyond quantitative improvements, DanceChat offers a new perspective on choreography modeling, where natural language serves not only as descriptive guidance but also as a bridge across modalities. 



\bibliography{aaai2026}

\begin{thebibliography}{43}
\providecommand{\natexlab}[1]{#1}

\bibitem[{Anthropic(2023)}]{anthropic2023claude}
Anthropic. 2023.
\newblock Claude 2 Model Card.

\bibitem[{Carroll and Moore(2008)}]{carroll2008feeling}
Carroll, N.; and Moore, M. 2008.
\newblock Feeling movement: Music and dance.
\newblock \emph{Revue internationale de philosophie}, (4): 413--435.

\bibitem[{Chan et~al.(2019)Chan, Ginosar, Zhou, and Efros}]{chan2019everybody}
Chan, C.; Ginosar, S.; Zhou, T.; and Efros, A.~A. 2019.
\newblock Everybody dance now.
\newblock In \emph{Proceedings of the IEEE/CVF international conference on computer vision}, 5933--5942.

\bibitem[{DeepSeek(2024)}]{deepseek2024}
DeepSeek. 2024.
\newblock DeepSeek-V2: Towards Language Agents with Internet-Scale Knowledge.

\bibitem[{Dhariwal et~al.(2020)Dhariwal, Jun, Payne, Kim, Radford, and Sutskever}]{dhariwal2020jukebox}
Dhariwal, P.; Jun, H.; Payne, C.; Kim, J.~W.; Radford, A.; and Sutskever, I. 2020.
\newblock Jukebox: A generative model for music.
\newblock \emph{arXiv preprint arXiv:2005.00341}.

\bibitem[{Ding et~al.(2024)Ding, Liu, Dong, Zhang, Qian, He, Lin, and Wang}]{ding2024songcomposer}
Ding, S.; Liu, Z.; Dong, X.; Zhang, P.; Qian, R.; He, C.; Lin, D.; and Wang, J. 2024.
\newblock Songcomposer: A large language model for lyric and melody composition in song generation.
\newblock \emph{arXiv preprint arXiv:2402.17645}.

\bibitem[{Doh et~al.(2024)Doh, Lee, Jeong, and Nam}]{doh2024enriching}
Doh, S.; Lee, M.; Jeong, D.; and Nam, J. 2024.
\newblock Enriching Music Descriptions with A Finetuned-LLM and Metadata for Text-to-Music Retrieval.
\newblock In \emph{ICASSP 2024-2024 IEEE International Conference on Acoustics, Speech and Signal Processing (ICASSP)}, 826--830. IEEE.

\bibitem[{Gardner et~al.(2024)Gardner, Durand, Stoller, and Bittner}]{gardner2023llark}
Gardner, J.; Durand, S.; Stoller, D.; and Bittner, R. 2024.
\newblock LLark: A Multimodal Instruction-Following Language Model for Music.
\newblock \emph{Proc. of the International Conference on Machine Learning (ICML)}.

\bibitem[{Guo et~al.(2022)Guo, Zou, Zuo, Wang, and Cheng}]{guo2022generating}
Guo, C.; Zou, S.; Zuo, X.; Wang, S.; and Cheng, L. 2022.
\newblock Generating Diverse and Natural 3D Human Motions from Text.
\newblock In \emph{Proceedings of the IEEE/CVF Conference on Computer Vision and Pattern Recognition (CVPR)}, 5152--5161.

\bibitem[{Han et~al.(2023)Han, Zhang, Shao, Gao, Xu, Xiao, Zhang, Liu, Wen, Guo et~al.}]{han2023imagebind}
Han, J.; Zhang, R.; Shao, W.; Gao, P.; Xu, P.; Xiao, H.; Zhang, K.; Liu, C.; Wen, S.; Guo, Z.; et~al. 2023.
\newblock Imagebind-llm: Multi-modality instruction tuning.
\newblock \emph{arXiv preprint arXiv:2309.03905}.

\bibitem[{Ho, Jain, and Abbeel(2020)}]{ho2020denoising}
Ho, J.; Jain, A.; and Abbeel, P. 2020.
\newblock Denoising diffusion probabilistic models.
\newblock \emph{Advances in neural information processing systems}, 33: 6840--6851.

\bibitem[{Huang et~al.(2021)Huang, Hu, Wu, Sawada, Zhang, and Jiang}]{DBLP:conf/iclr/HuangHWSZJ21}
Huang, R.; Hu, H.; Wu, W.; Sawada, K.; Zhang, M.; and Jiang, D. 2021.
\newblock Dance Revolution: Long-Term Dance Generation with Music via Curriculum Learning.
\newblock In \emph{9th International Conference on Learning Representations, {ICLR} 2021, Virtual Event, Austria, May 3-7, 2021}. OpenReview.net.

\bibitem[{Huang et~al.(2024)Huang, Xu, Xu, Zhang, Zheng, Qin, and He}]{huang2024beat}
Huang, Z.; Xu, X.; Xu, C.; Zhang, H.; Zheng, C.; Qin, J.; and He, S. 2024.
\newblock Beat-It: Beat-Synchronized Multi-Condition 3D Dance Generation.
\newblock In \emph{European Conference on Computer Vision}, 273--290. Springer.

\bibitem[{Kim, Kim, and Choi(2023)}]{kim2023flame}
Kim, J.; Kim, J.; and Choi, S. 2023.
\newblock Flame: Free-form language-based motion synthesis \& editing.
\newblock In \emph{Proceedings of the AAAI Conference on Artificial Intelligence}, volume~37, 8255--8263.

\bibitem[{Lee et~al.(2019)Lee, Yang, Liu, Wang, Lu, Yang, and Kautz}]{lee2019dancing}
Lee, H.-Y.; Yang, X.; Liu, M.-Y.; Wang, T.-C.; Lu, Y.-D.; Yang, M.-H.; and Kautz, J. 2019.
\newblock Dancing to music.
\newblock \emph{Advances in neural information processing systems}, 32.

\bibitem[{Li et~al.(2021)Li, Yang, Ross, and Kanazawa}]{li2021ai}
Li, R.; Yang, S.; Ross, D.~A.; and Kanazawa, A. 2021.
\newblock Ai choreographer: Music conditioned 3d dance generation with aist++.
\newblock In \emph{Proceedings of the IEEE/CVF International Conference on Computer Vision}, 13401--13412.

\bibitem[{Li et~al.(2023)Li, Zhao, Zhang, Su, Ren, Zhang, Tang, and Li}]{li2023finedance}
Li, R.; Zhao, J.; Zhang, Y.; Su, M.; Ren, Z.; Zhang, H.; Tang, Y.; and Li, X. 2023.
\newblock Finedance: A fine-grained choreography dataset for 3d full body dance generation.
\newblock In \emph{Proceedings of the IEEE/CVF International Conference on Computer Vision}, 10234--10243.

\bibitem[{Liang et~al.(2024)Liang, Zhang, Li, Yu, and Xu}]{liang2024intergen}
Liang, H.; Zhang, W.; Li, W.; Yu, J.; and Xu, L. 2024.
\newblock Intergen: Diffusion-based multi-human motion generation under complex interactions.
\newblock \emph{International Journal of Computer Vision}, 132(9): 3463--3483.

\bibitem[{Liu et~al.(2021)Liu, Shen, Zhang, Dolan, Carin, and Chen}]{liu2021makes}
Liu, J.; Shen, D.; Zhang, Y.; Dolan, B.; Carin, L.; and Chen, W. 2021.
\newblock What Makes Good In-Context Examples for GPT-$3 $?
\newblock \emph{arXiv preprint arXiv:2101.06804}.

\bibitem[{Liu and Chilton(2022)}]{liu2022design}
Liu, V.; and Chilton, L.~B. 2022.
\newblock Design guidelines for prompt engineering text-to-image generative models.
\newblock In \emph{Proceedings of the 2022 CHI conference on human factors in computing systems}, 1--23.

\bibitem[{Loper et~al.(2023)Loper, Mahmood, Romero, Pons-Moll, and Black}]{loper2023smpl}
Loper, M.; Mahmood, N.; Romero, J.; Pons-Moll, G.; and Black, M.~J. 2023.
\newblock SMPL: A skinned multi-person linear model.
\newblock In \emph{Seminal Graphics Papers: Pushing the Boundaries, Volume 2}, 851--866.

\bibitem[{McFee et~al.(2015)McFee, Raffel, Liang, Ellis, McVicar, Battenberg, and Nieto}]{mcfee2015librosa}
McFee, B.; Raffel, C.; Liang, D.; Ellis, D.~P.; McVicar, M.; Battenberg, E.; and Nieto, O. 2015.
\newblock librosa: Audio and music signal analysis in python.
\newblock In \emph{Proceedings of the 14th python in science conference}, volume~8, 18--25.

\bibitem[{Melechovsky et~al.(2023)Melechovsky, Guo, Ghosal, Majumder, Herremans, and Poria}]{melechovsky2023mustango}
Melechovsky, J.; Guo, Z.; Ghosal, D.; Majumder, N.; Herremans, D.; and Poria, S. 2023.
\newblock Mustango: Toward controllable text-to-music generation.
\newblock \emph{arXiv preprint arXiv:2311.08355}.

\bibitem[{OpenAI(2023)}]{openai2023gpt4}
OpenAI. 2023.
\newblock GPT-4 Technical Report.
\newblock arXiv:2303.08774.

\bibitem[{OpenAI API()}]{chatgptapi}
OpenAI API. 2024.
\newblock Https://openai.com/api/.

\bibitem[{Oppenlaender(2023)}]{oppenlaender2023taxonomy}
Oppenlaender, J. 2023.
\newblock A taxonomy of prompt modifiers for text-to-image generation.
\newblock \emph{Behaviour \& Information Technology}, 1--14.

\bibitem[{Radford(2018)}]{radford2018improving}
Radford, A. 2018.
\newblock Improving language understanding by generative pre-training.

\bibitem[{Radford et~al.(2021)Radford, Kim, Hallacy, Ramesh, Goh, Agarwal, Sastry, Askell, Mishkin, Clark et~al.}]{radford2021learning}
Radford, A.; Kim, J.~W.; Hallacy, C.; Ramesh, A.; Goh, G.; Agarwal, S.; Sastry, G.; Askell, A.; Mishkin, P.; Clark, J.; et~al. 2021.
\newblock Learning transferable visual models from natural language supervision.
\newblock In \emph{International conference on machine learning}, 8748--8763. PMLR.

\bibitem[{Ribeiro-Gomes et~al.(2024)Ribeiro-Gomes, Cai, Milacski, Wu, Prakash, Takagi, Aubel, Kim, Bernardino, and De~La~Torre}]{ribeiro2024motiongpt}
Ribeiro-Gomes, J.; Cai, T.; Milacski, Z.~{\'A}.; Wu, C.; Prakash, A.; Takagi, S.; Aubel, A.; Kim, D.; Bernardino, A.; and De~La~Torre, F. 2024.
\newblock MotionGPT: Human Motion Synthesis with Improved Diversity and Realism via GPT-3 Prompting.
\newblock In \emph{Proceedings of the IEEE/CVF Winter Conference on Applications of Computer Vision}, 5070--5080.

\bibitem[{Siyao et~al.(2022)Siyao, Yu, Gu, Lin, Wang, Qian, Loy, and Liu}]{siyao2022bailando}
Siyao, L.; Yu, W.; Gu, T.; Lin, C.; Wang, Q.; Qian, C.; Loy, C.~C.; and Liu, Z. 2022.
\newblock Bailando: 3d dance generation by actor-critic gpt with choreographic memory.
\newblock In \emph{Proceedings of the IEEE/CVF Conference on Computer Vision and Pattern Recognition}, 11050--11059.

\bibitem[{Tevet et~al.(2022)Tevet, Raab, Gordon, Shafir, Cohen-Or, and Bermano}]{tevet2022human}
Tevet, G.; Raab, S.; Gordon, B.; Shafir, Y.; Cohen-Or, D.; and Bermano, A.~H. 2022.
\newblock Human motion diffusion model.
\newblock \emph{arXiv preprint arXiv:2209.14916}.

\bibitem[{Touvron et~al.(2023)}]{touvron2023llama}
Touvron, H.; et~al. 2023.
\newblock LLaMA: Open and Efficient Foundation Language Models.
\newblock arXiv:2307.09288.

\bibitem[{Tseng, Castellon, and Liu(2023)}]{tseng2023edge}
Tseng, J.; Castellon, R.; and Liu, K. 2023.
\newblock Edge: Editable dance generation from music.
\newblock In \emph{Proceedings of the IEEE/CVF Conference on Computer Vision and Pattern Recognition}, 448--458.

\bibitem[{Van Den~Oord, Vinyals et~al.(2017)}]{van2017neural}
Van Den~Oord, A.; Vinyals, O.; et~al. 2017.
\newblock Neural discrete representation learning.
\newblock \emph{Advances in neural information processing systems}, 30.

\bibitem[{Wang et~al.(2024)Wang, Liu, Zhang, and Xu}]{motionagent}
Wang, H.; Liu, J.; Zhang, L.; and Xu, K. 2024.
\newblock Motion-Agent: A Conversational Framework for Human Motion Generation with LLMs.
\newblock \emph{arXiv preprint arXiv:2402.12345}.

\bibitem[{Wu et~al.(2023)Wu, Fei, Qu, Ji, and Chua}]{wu2023next}
Wu, S.; Fei, H.; Qu, L.; Ji, W.; and Chua, T.-S. 2023.
\newblock Next-gpt: Any-to-any multimodal llm.
\newblock \emph{arXiv preprint arXiv:2309.05519}.

\bibitem[{Yang, Yang, and Wang(2023)}]{yang2023longdancediff}
Yang, S.; Yang, Z.; and Wang, Z. 2023.
\newblock Longdancediff: Long-term dance generation with conditional diffusion model.
\newblock \emph{arXiv preprint arXiv:2308.11945}.

\bibitem[{Zhang et~al.(2024{\natexlab{a}})Zhang, Tang, Zhang, Lin, Han, Xiao, and Wang}]{zhang2024bidirectional}
Zhang, C.; Tang, Y.; Zhang, N.; Lin, R.-S.; Han, M.; Xiao, J.; and Wang, S. 2024{\natexlab{a}}.
\newblock Bidirectional Autoregessive Diffusion Model for Dance Generation.
\newblock In \emph{Proceedings of the IEEE/CVF Conference on Computer Vision and Pattern Recognition}, 687--696.

\bibitem[{Zhang et~al.(2024{\natexlab{b}})Zhang, Cai, Pan, Hong, Guo, Yang, and Liu}]{zhang2024motiondiffuse}
Zhang, M.; Cai, Z.; Pan, L.; Hong, F.; Guo, X.; Yang, L.; and Liu, Z. 2024{\natexlab{b}}.
\newblock Motiondiffuse: Text-driven human motion generation with diffusion model.
\newblock \emph{IEEE Transactions on Pattern Analysis and Machine Intelligence}.

\bibitem[{Zhou et~al.(2024)Zhou, Dou, Cao, Liao, Wang, Wang, Liu, Komura, Wang, and Liu}]{zhou2024emdm}
Zhou, W.; Dou, Z.; Cao, Z.; Liao, Z.; Wang, J.; Wang, W.; Liu, Y.; Komura, T.; Wang, W.; and Liu, L. 2024.
\newblock Emdm: Efficient motion diffusion model for fast and high-quality motion generation.
\newblock In \emph{European Conference on Computer Vision}, 18--38. Springer.

\bibitem[{Zhou et~al.(2019)Zhou, Barnes, Lu, Yang, and Li}]{zhou2019continuity}
Zhou, Y.; Barnes, C.; Lu, J.; Yang, J.; and Li, H. 2019.
\newblock On the continuity of rotation representations in neural networks.
\newblock In \emph{Proceedings of the IEEE/CVF conference on computer vision and pattern recognition}, 5745--5753.

\bibitem[{Zhou, Wan, and Wang(2024)}]{zhou2024avatargpt}
Zhou, Z.; Wan, Y.; and Wang, B. 2024.
\newblock AvatarGPT: All-in-One Framework for Motion Understanding Planning Generation and Beyond.
\newblock In \emph{Proceedings of the IEEE/CVF Conference on Computer Vision and Pattern Recognition}, 1357--1366.

\bibitem[{Zhuang et~al.(2022)Zhuang, Wang, Chai, Wang, Shao, and Xia}]{zhuang2022music2dance}
Zhuang, W.; Wang, C.; Chai, J.; Wang, Y.; Shao, M.; and Xia, S. 2022.
\newblock Music2dance: Dancenet for music-driven dance generation.
\newblock \emph{ACM Transactions on Multimedia Computing, Communications, and Applications (TOMM)}, 18(2): 1--21.

\end{thebibliography}


\end{document}